\documentclass{article}

\usepackage{microtype}
\usepackage{graphicx}
\usepackage{subfigure}
\usepackage{booktabs} 
\usepackage{makecell}
\usepackage{tikz}
\usetikzlibrary{calc} 

\usepackage{hyperref}

\usepackage[accepted]{icml_fmwild}

\usepackage{amsmath}
\usepackage{amssymb}
\usepackage{mathtools}
\usepackage{amsthm}

\usepackage[capitalize,noabbrev]{cleveref}

\theoremstyle{plain}

\theoremstyle{definition}

\theoremstyle{remark}

\usepackage[textsize=tiny]{todonotes}

\renewcommand{\paragraph}[2][.]{\vspace{0.1pt}\noindent{\bf #2#1}}

\icmltitlerunning{Domain-Aware Fine-Tuning of Foundation Models}

\begin{document}

\twocolumn[
\icmltitle{Domain-Aware Fine-Tuning of Foundation Models}



\icmlsetsymbol{equal}{*}

\begin{icmlauthorlist}
\icmlauthor{U\v gur Ali Kaplan\textsuperscript{$\dagger$}}{ut}
\icmlauthor{Margret Keuper}{mannheim,mpi}
\icmlauthor{Anna Khoreva}{bosch}
\icmlauthor{Dan Zhang}{ut,bosch}
\icmlauthor{Yumeng Li}{mannheim,bosch}
\end{icmlauthorlist}
\icmlaffiliation{bosch}{Bosch Center for Artificial Intelligence}
\icmlaffiliation{ut}{University of Tübingen}
\icmlaffiliation{mannheim}{University of Mannheim}
\icmlaffiliation{mpi}{Max Planck Institute for Informatics}
\icmlcorrespondingauthor{Yumeng Li}{yumeng.li@de.bosch.com}


\icmlkeywords{Foundation Models, Vision-Language Models, Finetuning, Domain Generalization}

\vskip 0.3in
]



\printAffiliationsAndNotice{\textsuperscript{$\dagger$} Work done at Bosch Center for AI}  

\begin{abstract}
Foundation models (FMs) have revolutionized computer vision, enabling effective learning across different domains. However, their performance under domain shift is yet underexplored. This paper investigates the zero-shot domain adaptation potential of FMs by comparing different backbone architectures and introducing novel domain-aware components that leverage domain related textual embeddings. 
We propose \textbf{dom}a\textbf{i}n adaptive \textbf{no}rmalization, termed as \textbf{Domino}, which explicitly leverages domain embeddings
during fine-tuning, thus making the model domain aware.
Ultimately, Domino enables more robust computer vision models that can adapt effectively to various unseen domains.\end{abstract}
\section{Introduction}\label{sec:intro}

Computer vision has made tremendous progress in recent years, thanks to the development of powerful neural network architectures and large-scale datasets \cite{deng2009imagenet, simonyan2014very, long2015fully, ronneberger2015u, hao2020brief, dosovitskiy2020image, li2022exploring, schuhmann2022laion}. However, when faced with domain shift - a common problem in real-world applications where the target domain has different characteristics than the training data - the performance of these models drops significantly. This is particularly problematic for tasks that require zero-shot domain adaptation, where during training there is no sample available in the target domain, though we might have an estimation of the potential domains \cite{farahani2021brief,liu2022deep}.


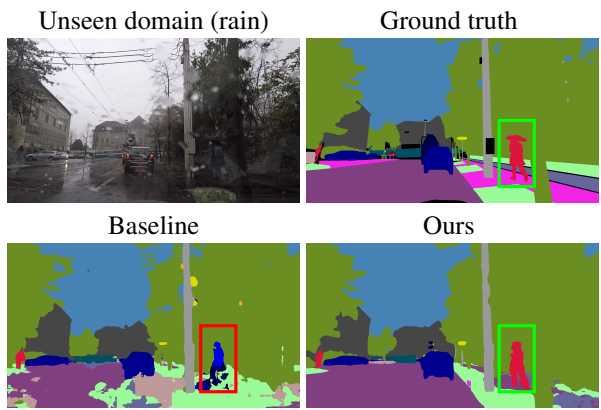
\begin{figure}[t]
    \begin{centering}
    \setlength{\tabcolsep}{0.0em}
    \renewcommand{\arraystretch}{0}
    \par\end{centering}
    \begin{centering}
    \vspace{-0.5em}
    \hfill{}%
	\begin{tabular}{@{}c@{}c}
        \centering
		Unseen domain (rain) & Ground truth \tabularnewline
	    \includegraphics[width=0.47\linewidth]{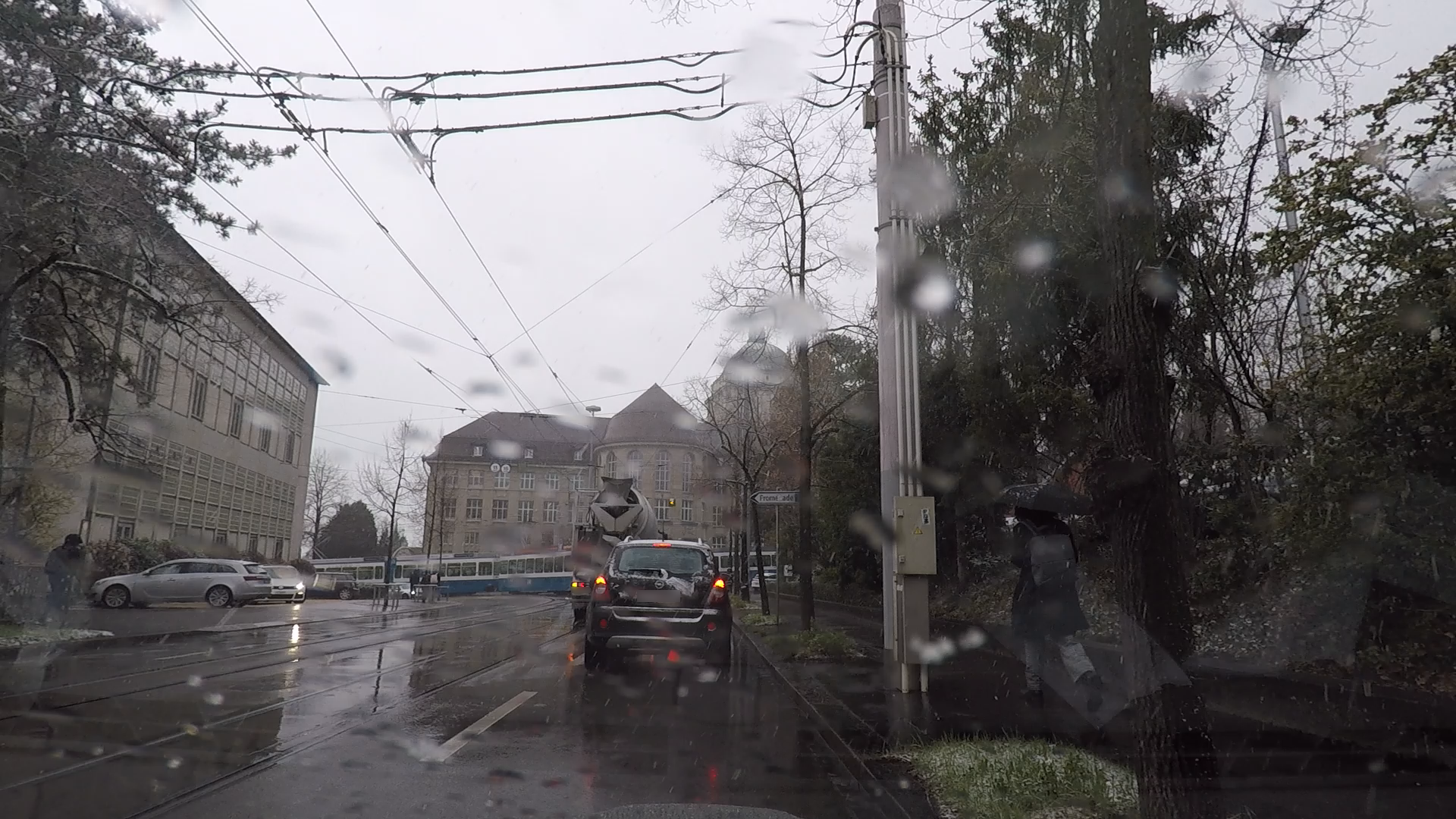} 
		&{\footnotesize{}} 
		\begin{tikzpicture}
            \node [
	        above right,
	        inner sep=0] (image) at (0,0) {\includegraphics[width=0.47\linewidth]{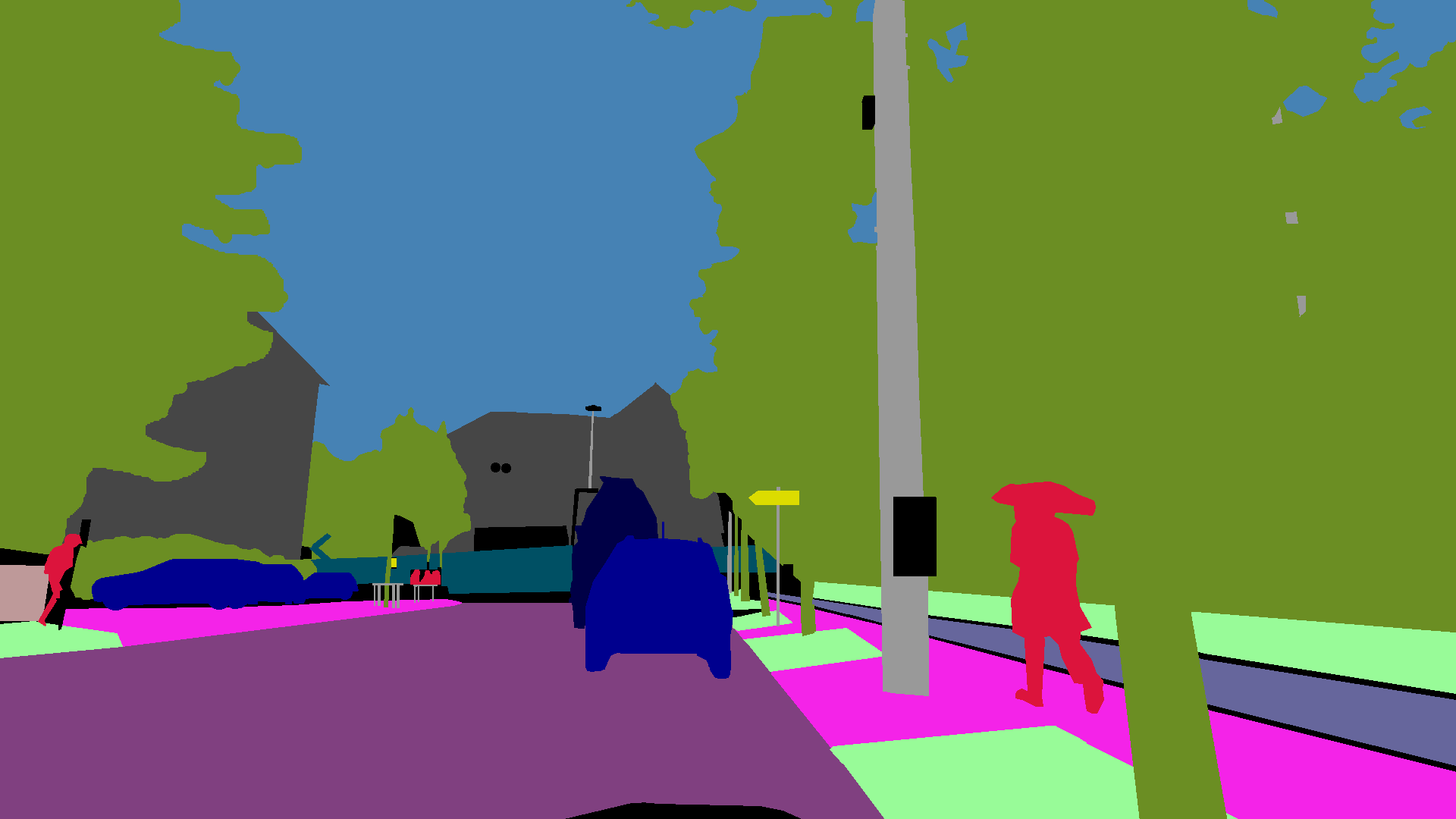} };
            \begin{scope}[
            x={($0.1*(image.south east)$)},
            y={($0.1*(image.north west)$)}]
            \draw[very thick,green] (6.6, 1.0) rectangle (7.8, 5);
        \end{scope}
    \end{tikzpicture}
		\tabularnewline
		Baseline & Ours \tabularnewline
	\begin{tikzpicture}
            \node [
	        above right,
	        inner sep=0] (image) at (0,0) {\includegraphics[width=0.47\linewidth]{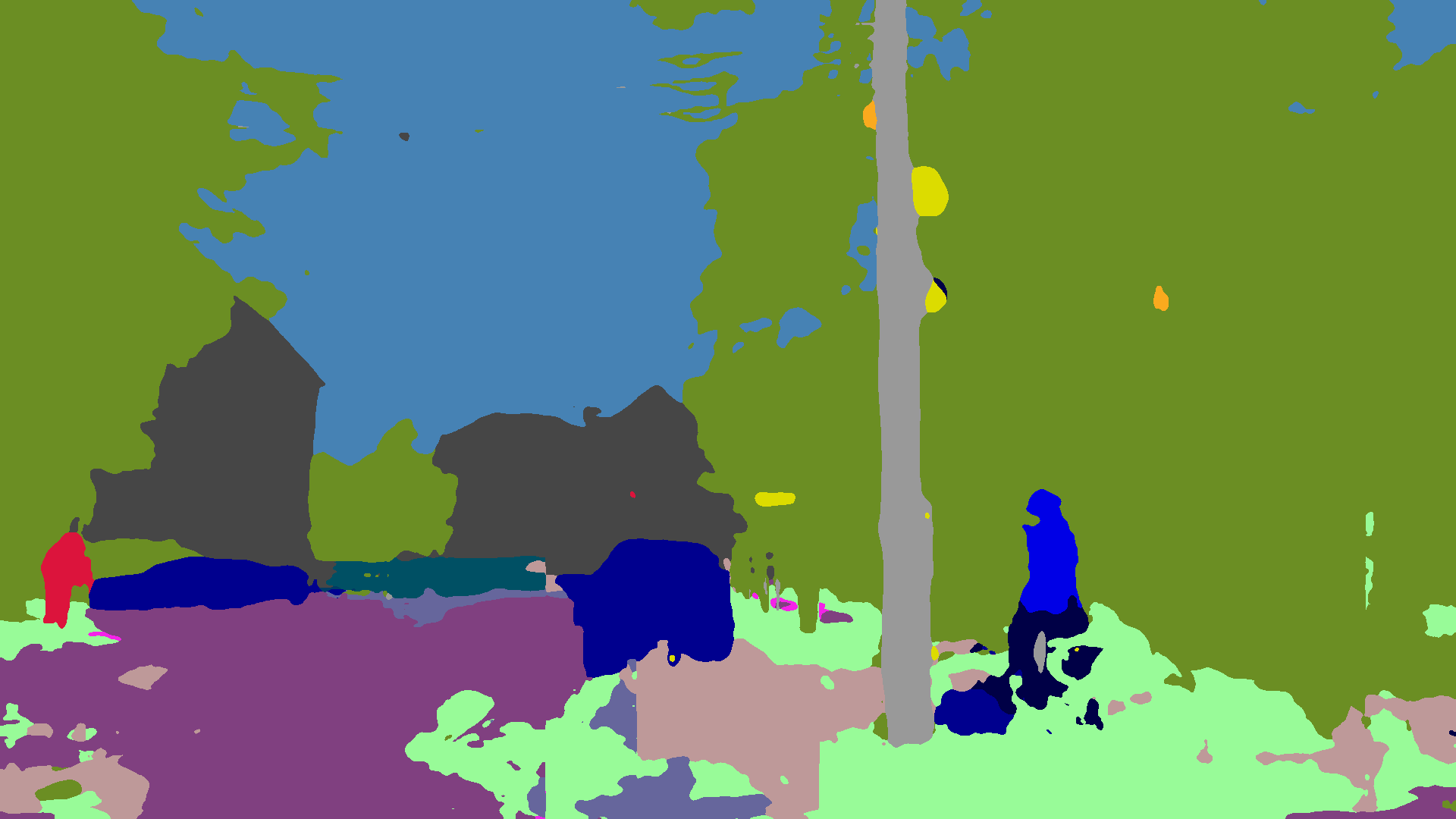}};
            \begin{scope}[
            x={($0.1*(image.south east)$)},
            y={($0.1*(image.north west)$)}]
            \draw[very thick,red] (6.6, 1.0) rectangle (7.8, 5);
        \end{scope}
    \end{tikzpicture} 
		&{\footnotesize{}} 
    \begin{tikzpicture}
            \node [
	        above right,
	        inner sep=0] (image) at (0,0) {\includegraphics[width=0.47\linewidth]{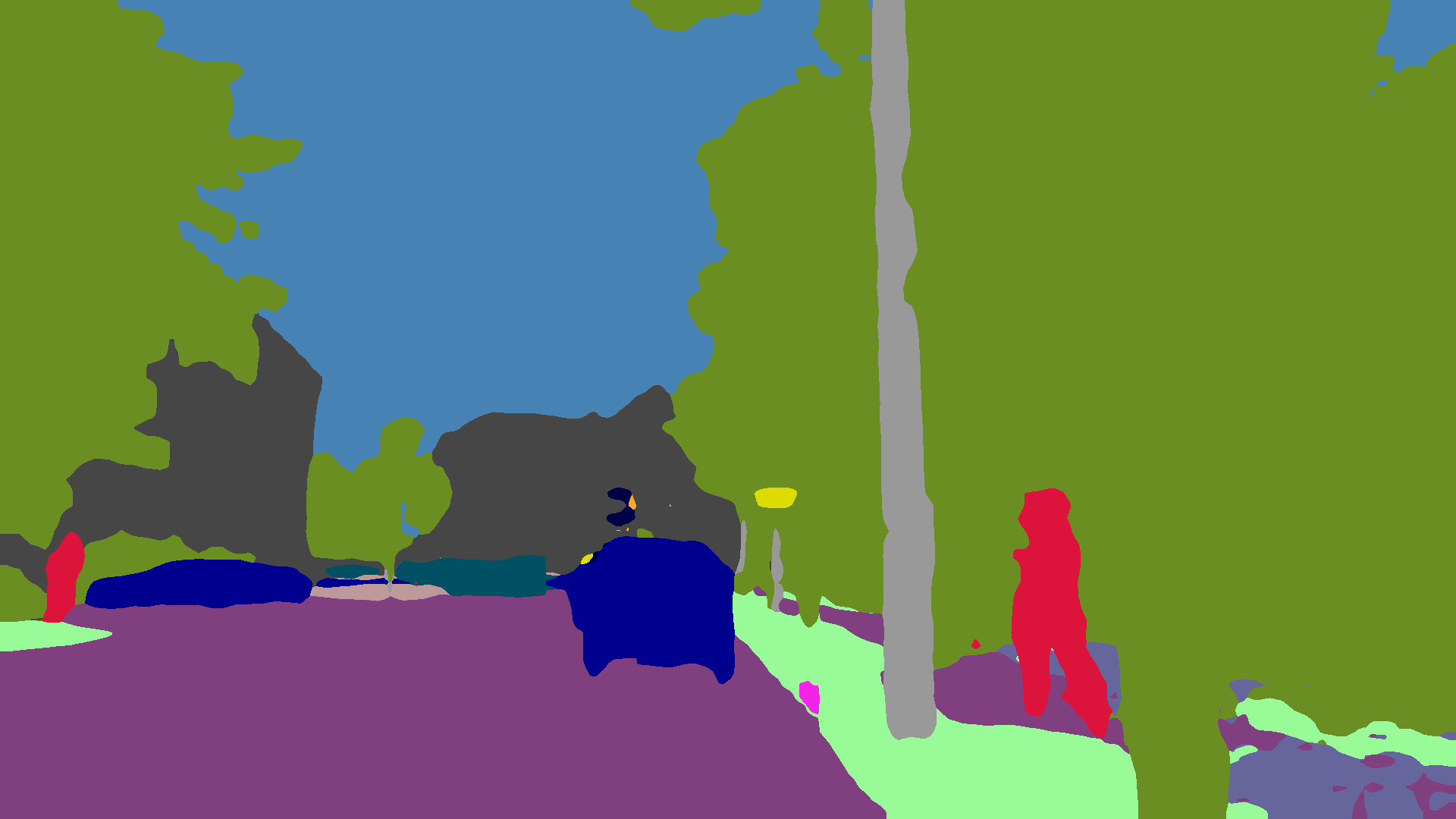}};
            \begin{scope}[
            x={($0.1*(image.south east)$)},
            y={($0.1*(image.north west)$)}]
           \draw[very thick,green] (6.6, 1.0) rectangle (7.8, 5);
        \end{scope}
    \end{tikzpicture}	
		\end{tabular}
\hfill{}
\par\end{centering}
\caption{
Traditional segmentation models trained on Cityscapes~\cite{Cordts2016Cityscapes} fail to make robust predictions under adverse weather conditions~\cite{SDV21}, where the person is misclassified. By employing the proposed domain-aware FM fine-tuning, our model can successfully detect the person, potentially avoiding accidents. 
}
\label{fig:intro-semseg}
\vspace{-1.7em}
\end{figure}

Foundation models (FMs), have become pivotal in various applications, from natural language processing to computer vision. Due to the large-scale pretraining, FMs have generalizable representations. Despite showing promising results on zero-shot classification tasks \cite{brown2020language, oquab2023dinov2}, it is yet challenging to employ these models directly for dense prediction tasks, e.g., semantic segmentation (see 1\textsuperscript{st} row in \cref{tab:domain-results}), necessitating the need of fine-tuning for the specific task. However, this will inevitably lead to knowledge forgetting and raise the question of how robust the adapted models are under domain shifts. Alternatively, \citet{gong2023prompting} proposed a prompt tuning method for test-time model adaptation.

To investigate the zero-shot domain adaptation performance of FMs, we first compare different FM backbone architectures on the challenging semantic segmentation task. We evaluate various vision backbones in \cref{tab:backbones}, such as DINOv2 \cite{oquab2023dinov2, darcet2023vision}, ResNet-50 and ResNetRS-420 \cite{he2016deep, bello2021revisiting}, CLIP-based fine-tuning methods leveraging MaskCLIP \cite{dong2023maskclip}, and Stable Diffusion (SD) \cite{rombach2021highresolution} based fine-tuning, i.e., VPD \cite{zhao2023unleashing}.
We observe that these models are generally not robust to domain shifts, where there is a considerable performance drop when tested on unseen domains. 

We hypothesized that visual embeddings can vary considerably with different domains, making the model vulnerable to changes such as weather and lighting conditions. To mitigate this issue, we propose to incorporate textual domain embeddings and introduce \textbf{dom}a\textbf{i}n adaptive \textbf{no}rmalization, termed as \textbf{Domino} during fine-tuning.
We employed CLIP \cite{radford2021learning} to automatically extract the domain embedding, eliminating the need for manual definition of the domain concept. 
Further, we explicitly leverage the domain embedding through Domino, making the model domain aware. Domino can effectively improve the model's generalization ability across different domains without requiring data from the target domain during training (see  \cref{fig:intro-semseg}).

Among all FM backbones, we observe that Stable Diffusion delivers the best generalization performance. Therefore, we employ it as the default backbone to test the proposed Domino, where we observed a significant enhancement on the model generalization capability with domain-aware fine-tuning. 
Meanwhile, SD can not only serve as the backbone, its generative nature benefits us to explore diverse synthetic data potentially resembling the target domain, which helps to close the domain gap. We discovered that with a proper mixing ratio of real and synthesized data (see \cref{tab:synth-results}), the model's generalization performance can be further boosted.

In summary, we empirically study the generalization capability of different FM backbones. Further, we propose a domain-aware fine-tuning strategy to explicitly leverage the domain information, leading to more robust models, especially when combined with diverse synthetic data.

\vspace{-0.8em}
\section{Method}\label{sec:method}
\vspace{-0.2em}
\setlength{\abovedisplayskip}{6.0pt}
\setlength{\belowdisplayskip}{5.5pt}

\subsection{Preliminary}
Stable Diffusion has demonstrated astonishing text-to-image synthesis capability, thanks to their large-scale pretraining. Naturally, it has learned rich multi-modal representations. 
Recent work VPD \cite{zhao2023unleashing} has explored its potential for downstream applications, e.g., depth estimation and semantic segmentation.
More specifically, they finetune the denoising UNet and extract intermediate features and cross-attention maps from it, which can provide semantically meaningful features \cite{hertz2022prompt,li2023divide}. Note that semantic classes are used as prompts for semantic segmentation task, and a text adapter is introduced as well. Further, a lightweight task-specific decoder taking extracted features is incorporated.

Despite showing promising results on the in-domain evaluation, it remains unclear how this model can perform under domain shift. In other words, it's not yet explored how the prior knowledge of powerful Stable Diffusion can help the downstream applications, which is of greater interest, and the focus of this work. Additionally, we propose a novel domain-aware fine-tuning strategy, where we extract the domain embeddings with CLIP (see  \cref{subsec:domain-extraction}), and incorporate them into the fine-tuning pipe to enhance domain awareness (see \cref{subsec:dede}).

\begin{figure}[t]
\begin{center}
\centering
\includegraphics[width=0.95\linewidth]{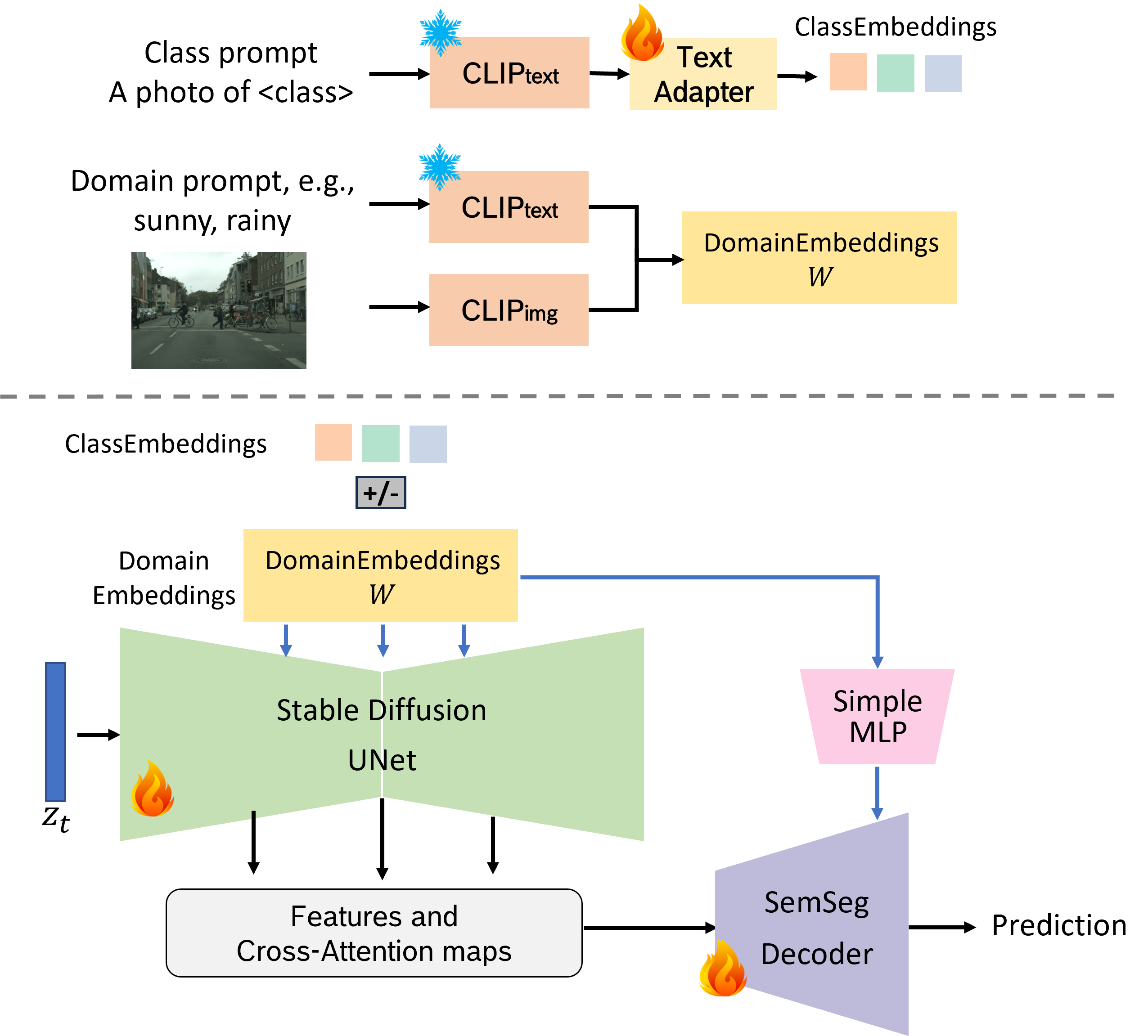}
\vspace{-1.0em}
\caption{Trainable CLIP image encoder integrated into VPD architecture. Domain embedding is computed from input image and added/subtracted from text adapter output to manipulate domain cues. Cross-attention calculation with strengthened/weakened cues improves segmentation performance in both in-domain and domain shift scenarios.}
\label{fig:vpd-spade}
\end{center}
\vspace{-1.5em}
\end{figure}

\subsection{Automatic Domain Embedding Extraction}\label{subsec:domain-extraction}
The domain concept is quite often assigned manually. 
Instead, we seek a way to automatically obtain the domain information. Prior work \cite{wang2023exploring} has shown that CLIP \cite{radford2021learning} is capable of assessing the look and feel of images. Thus, we propose to leverage CLIP to automatically extract the domain embeddings, which can be further utilized to enhance domain-awareness during FM fine-tuning.

\begin{table*}[t]
\caption{Comparison of the different foundation models as backbone for semantic segmentation.}
\label{tab:backbones}
\vskip 0.08in
\begin{center}
\begin{small}
\begin{sc}
\begin{tabular}{lccccccr}
\hline
\textbf{Backbone} & \textbf{Pre-Training} & \textbf{Parameter Size} & \makecell{\textbf{Cityscapes} \\ \textbf{mIoU}  $\uparrow$} & \makecell{\textbf{ACDC} \\ \textbf{mIoU}  $\uparrow$} & \textbf{mIoU\%}  $\uparrow$\\
\hline
ViT-L/14 & DINOv2 with Registers & 304M & 49.90 & 22.35 & 47.79 \\
ViT-B/16 & CLIP (MaskCLIP+)  & 147M & 73.51 & 36.98 & 50.31 \\
ResNet-101 & CLIP (MaskCLIP+) & 156M & 73.95 & 35.27 & 47.69 \\
ResNet-50 & Supervised & 28.5M & 69.68 & 38.28 & 54.94 \\
ResNetRS-420 UNet & Supervised & 205M & 73.74 & 50.85 & 68.96 \\
VPD & Stable Diffusion & 868M & \textbf{76.82} & \textbf{62.50} & \textbf{81.36} \\\hline
\end{tabular}
\vspace{-0.5em}
\end{sc}
\end{small}
\end{center}
\vspace{-1em}
\end{table*}

To calculate the domain embedding, we begin by defining base cases that describe the aspects of interest. For instance, in autonomous driving, related domains involve different weather conditions and time of the day.
Each of these domain descriptions are encoded using the CLIP text encoder:
\vspace{-0.5em}
\begin{align*}
d_i &= \text{CLIP}_{\text{Text}}(k_i)\ \forall i = 1, \ldots, N\,,
\end{align*}
where $K = \{k_1, \ldots, k_N \}$ are defined domain descriptions and $D = \{d_1, \ldots, d_N\}$ is the resulting description embeddings. Given an image, we can obtain the image embedding through the CLIP image encoder and compute the similarity with each individual description embedding. Finally, the domain embedding of the given image is computed via a weighted sum of all description embeddings: 
\begin{align*}
    I =& \ \text{CLIP}_{\text{Image}}(x) \\
    \alpha_i =& \ \text{Softmax}\left (  \frac{I \cdot d_i}{||I|| \cdot ||d_i||} \right), \\
    W =&  \sum_i \alpha_i \cdot d_i\, ,
\end{align*} 
where $x$ represents the input image, $\alpha_i$ and $W$ denotes the similarity and final domain embedding, respectively.
By doing so, we eliminate the need for manual domain assignment and mitigate the ill-defined one-to-one mapping issue by taking a weighted average.

\subsection{Domain Aware Fine-tuning}\label{subsec:dede}
Stable Diffusion naturally is capable of utilizing textual embedding, and thus we can simply combine the domain embedding with the original prompt embedding, i.e., class embedding, as illustrated in \cref{fig:vpd-spade}.
There are two possible ways of combination, namely addition or subtraction. When adding the domain embedding, we encourage the model to explore the domain information for prediction. While for subtraction, we essentially try to remove domain-related information, thus enforcing domain-invariant learning. As experimented in \cref{tab:domain-results}, we found the latter is more effective in improving the generalization performance.

In addition to being utilized by the Stable Diffusion backbone, we propose to incorporate the domain embedding in the segmentation decoder head to further enhance domain awareness. 
Inspired by SPADE \cite{park2019SPADE}, we introduce \textbf{dom}a\textbf{i}n adaptive \textbf{no}rmalization, termed as \textbf{Domino}. Specifically, we map the domain embedding through simple MLP layers into modulation parameters $\gamma$ and $\beta$.
\begin{align}\label{eq:modulation}
 f_{adp} = \frac{f - \mu_{f}}{\gamma_{f}} \gamma(W) + \mu(W),
\end{align}
where $f$ are intermediate features from the segmentation decoder,  $\mu_{f}$ and $\gamma_{f}$ denote their mean and standard deviation. $\mu(W)$ and $\gamma(W)$ are the learned affine transformation parameters, conditioned on the domain embedding $W$. 
In doing so, we effectively inject domain information into the segmentation decoder, making it domain-aware.

\vspace{-0.5em}
\section{Experiments}\label{sec:exp}
\vspace{-0.4em}

\paragraph{Experimental Setup}
In this study, we focus on the challenging task of semantic segmentation, which requires per-pixel semantic class prediction.
We train the models on the Cityscapes \cite{Cordts2016Cityscapes} training set, and evaluate the model's generalization performance on ACDC \cite{SDV21}.
Cityscapes is an urban scene dataset with 19 semantic classes, which is collected mainly in Germany under good weather conditions during daytime.
In contrast, ACDC contains more adverse weather conditions, such as 
fog, snow, rain and nighttime. Compared to Cityscapes, there is a significant domain shift,  making it a challenging test dataset meanwhile a perfect testbed for assessing the model's generalization capability.

For evaluation, we use the mean intersection-over-union (mIoU) \cite{everingham2015pascal}. 
Similar to \cite{fahes2023poda}, we also report the domain adaptation performance as the percentage of target domain mIoU (ACDC) over source domain mIoU (Cityscapes):
\begin{align*}
    \text{mIoU}\% = 100 \times \frac{\text{ACDC mIoU}}{\text{Cityscapes mIoU}}
\end{align*}
We train all models on A100 GPUs using single-GPU training, using weighted cross-entropy loss. 
For a fair comparison, we train all models for 80,000 iterations. We use the AdamW optimizer \cite{loshchilov2018decoupled} with a learning rate of $8e-5$ and weight decay $1e-3$. Diffusion transformer is using a different learning rate of $0.1$ while the text encoder is frozen as in the VPD paper. We use poly learning rate scheduling to decrease the learning rate to 0 as training progresses.

\paragraph{Comparison of FM Backbones}
In this comparison, we include a wide range of FMs, such as self-supervised DINOv2 \cite{oquab2023dinov2, darcet2023vision}, CLIP \cite{radford2021learning} and Stable Diffusion \cite{rombach2021highresolution}.

Additionally, to provide a comparison against a similar model structure, we combine pre-trained ResNetRS-420's \cite{bello2021revisiting} in UNet form and fine-tune it on the Cityscapes dataset. We use CLIP-based methods with ViT and ResNet backbones, based on MaskCLIP+ \cite{dong2023maskclip}.

In Table \ref{tab:backbones}, we present the evaluation results of all models finetuned with cross entropy loss. 
We can see that VPD (built upon Stable Diffusion) performs the best on the  Cityscapes validation set, as well as on the unseen ACDC, which presents a considerable domain shift. Notably, CLIP backbones and supervised models come close to the VPD performance on Cityscapes, but all of them show a significant performance degradation under domain shift. A surprising observation is the evident under-performance of the DINOv2 model. We attribute this to its need for a longer training time. In our experiments, we observed that while the validation performance of the other models kept increasing slowly nearing the end of our training schedule, DINOv2's validation scores kept increasing quickly, and it could have benefited from more training time. However, for a fair comparison, we have trained it using the same training schedule as the other models, which led to its underfitting on Cityscapes.

\begin{table}[t]
\caption{
Comparison of architectural modifications.
Using a frozen backbone leads to the worst results. Our proposed Domino can effectively boost the generalization performance, especially when subtracting the domain information to encourage domain-invariant learning.
}
\label{tab:domain-results}
\vskip 0.15in
\begin{center}
\begin{small}
\begin{sc}
\begin{tabular}{lcccr}
\toprule
\textbf{Variant} & \makecell{\textbf{Cityscapes} \\ \textbf{mIoU}  $\uparrow$} & \makecell{\textbf{ACDC} \\ \textbf{mIoU}$ \uparrow$} & \textbf{mIoU\%} $\uparrow$  \\
\midrule
VPD (Frozen) & 69.19 & 50.48 & 72.96 \\
VPD & 76.87 & 62.61 & 81.45 \\
Domino-Add & \textbf{77.20} & 63.53 & 82.29 \\
Domino-Sub & 76.13 & \textbf{65.00} & \textbf{85.38} \\
\bottomrule
\end{tabular}
\vspace{-0.8em}
\end{sc}
\end{small}
\end{center}
\vspace{-1.0em}
\end{table}

\begin{table}[t]
\caption{Comparison of varying mixing ratio of real and synthetic data.  An even split provides the best generalization performance.}
\label{tab:synth-results}
\vskip 0.05in
\begin{center}
\begin{small}
\begin{sc}
\begin{tabular}{cccc}
\toprule
\makecell{\textbf{Real}  / \\ \textbf{Synthetic} } &  \makecell{\textbf{Cityscapes} \\ \textbf{mIoU}  $\uparrow$} & \makecell{\textbf{ACDC} \\ \textbf{mIoU}  $\uparrow$}& \textbf{mIoU\%}  $\uparrow$ \\
\midrule
100/0 & 76.87 & 62.61 & 81.45 \\
75/25 & \textbf{77.45} & 63.52 &  82.01\\
50/50 & 77.07 & \textbf{64.49} & \textbf{83.68} \\
0/100 & 71.24 & 56.84 & 79.79 \\
\bottomrule
\end{tabular}
\vspace{-1.0em}
\end{sc}
\end{small}
\end{center}
\vspace{-1em}
\end{table}

Another observation is that despite CLIP-based models also being pre-trained on large-scale image-text pairs, they have a substantial degradation on ACDC compared to Stable Diffusion.  
We hypothesize this is due to CLIP's tendency of catastrophic forgetting when fine-tuned fully, while being trained with a generative objective, Stable Diffusion is more tolerant in this regard.
Ultimately, this validates our selection of Stable Diffusion as the FM backbone for further improvement of zero-shot domain adaptation.

\paragraph{Effect of Domino}
In \cref{tab:domain-results}, we present the comparison of the proposed domain-aware fine-tuning Domino with the original VPD \cite{zhao2023unleashing}. 
We first examine the impact of architectural changes on performance, taking into account prior work suggesting that some foundational models perform better with frozen weights. To explore this phenomenon, we experimented using a frozen Stable Diffusion backbone (1\textsuperscript{st} row in \cref{tab:domain-results}), which has the worst performance. This indicates the necessity of fine-tuning the entire backbone for the specific downstream task. 
We implemented both domain addition and subtraction, denoted as Domino-Add and Domino-Sub in \cref{tab:domain-results}. We observe that both variants outperform the baseline VPD in generalization performance on ACDC, which indicates the effectiveness of domain-aware fine-tuning. Notably, the Domino-Sub version achieves the best generalization results. This suggests that by encouraging domain-invariant representation learning during training, the model becomes more robust under domain shifts. We also see that Domino-Add can improve in-domain performance. We hypothesize that encouraging the usage of domain embedding can ease pattern learning as Cityscapes mostly consist of clear day images.

\paragraph{Effect of Synthetic Data}
We investigate the impact of incorporating synthetic data into our framework for zero-shot domain adaptation. Prior work has demonstrated that utilizing synthetic data during training can enhance performance for traditional models \cite{beery2020synthetic, li2023issa,li2024aldm,wang2024lumen}. However, the effectiveness of synthetic data with Foundation Models (FMs) is a topic of discussion, particularly as these models have been trained on massive real data. Yet, we hypothesized for improving domain generalization, it is crucial for the model to see diverse data from different domains, e.g., under different weather conditions. To explore this, we employ ALDM \cite{li2024aldm} to generate synthetic data based on the labels of Cityscapes training set and diverse prompts.

In \cref{tab:synth-results}, we present the results of fine-tuning the base VPD model with varying ratios between real and synthetic data. Notably, combining synthetic data with real data leads to improved performance in both domains. As we increase the proportion of synthetic data, target domain performance also increases. However, when using synthetic data only, we observe a performance decrease in both domains. This
finding suggests that relying solely on synthetic data is not beneficial, as the quality of synthetic might not be as good as real ones. This also highlights the importance of incorporating real data alongside synthetic data for foundational models during fine-tuning.

\vspace{-0.5em}
\section{Conclusion \& Discussion}
\vspace{-0.5em}
In this study, we conducted extensive experiments comparing different foundation models as the backbone for zero-shot domain adaptation in semantic segmentation. We empirically observe that Stable Diffusion based VPD model achieves better generalization performance. 
We then demonstrated the proposed Domino with explicit usage of the domain information can significantly boost the model's generalization further. 
Our results indicate that domain embedding addition encourages the use of domain cues, which can be beneficial for improving the in-domain segmentation performance. In contrast, domain embedding subtraction encourages the use of more domain-invariant features which can enhance the generalization performance. Furthermore, we have found that incorporating synthetic data with a proper ratio during the foundational model fine-tuning is also beneficial for domain generalization performance.

\section*{Acknowledgement}
U\v gur Ali Kaplan would like to thank Prof. Matthias Hein and Dr. Nicole Ludwig for their discussions during the project.

\bibliography{reference}
\bibliographystyle{icml2024}


\end{document}